\documentclass[sigconf,nonacm]{acmart}

\renewcommand\footnotetextcopyrightpermission[1]{}
\usepackage{booktabs}
\usepackage{multirow}
\usepackage{amsmath}
\usepackage{graphicx}
\usepackage{xspace}
\usepackage{microtype}
\usepackage{placeins}

\newcommand{\method}{\textsc{MIS}\xspace}

\newcommand{\pp}{\,\mathrm{pp}}

\title{Towards Efficient Whole Page Ranking with One-Pass Inference}
\author{\parbox{\textwidth}{\centering
Qi Xu\textsuperscript{\textdagger} \quad Zhigang Hua\textsuperscript{\textdagger} \quad Zijie Lei \quad Yan Xie \quad Jizhe Zhang \quad Lynn Lu \quad Delun Du\\
Sha Hua \quad Ren Mao \quad Ke Pan \quad Haomin Yu \quad Sandeep Pandey \quad Luke Simon\textsuperscript{\textdagger}\\[6pt]}}

\affiliation{\institution{Meta Platforms, Inc.}\country{United States}}
\begin{document}

\begin{abstract}
Whole-page ranking jointly optimizes advertisements and organic content to balance business outcomes and user engagement under complex layout and business constraints. Existing non-autoregressive (NAR) approaches enable efficient page generation but typically score candidate items independently of previously selected items, limiting their ability to capture order-dependent interactions. Autoregressive (AR) methods address this limitation through sequential conditioning, but incur substantial inference overhead that challenges large-scale deployment. Bridging this quality--efficiency gap remains a challenge in industrial whole-page ranking.

We propose Multi-Item Scoring (\method), a prefix-aware NAR-style framework that incorporates sequential context into efficient multi-item scoring. MIS integrates global candidate interactions, localized prefix conditioning, and position-aware scoring within a shared representation architecture. By computing reusable representations once per request and performing lightweight prefix-conditioned scoring during constrained beam search, MIS captures important sequential dependencies without repeated full-model inference.

Experiments on KuaiRand-1K and RL4RS demonstrate that MIS approaches the generation quality of an autoregressive CGA baseline while achieving $4.07\times$ and $3.54\times$ faster CPU decoding, respectively. In a randomized evaluation on a large-scale social-video platform, MIS improves the realization metric by $0.104\%$ and watch time by $0.058\%$, while reducing ranking-service CPU consumption by $9.73\%$. These results demonstrate that prefix-aware multi-item scoring can narrow the quality gap between NAR and AR generation while retaining the efficiency required for industrial scalability.
\end{abstract}



\maketitle
\begingroup
\renewcommand{\thefootnote}{\textdagger}
\footnotetext{Corresponding author: Qi Xu (xuqi@meta.com), Zhigang Hua (zhua@meta.com), Luke Simon (lukesimon@meta.com)}
\endgroup
\pagestyle{plain} 

\section{Introduction}
Large-scale recommender systems typically retrieve and rank candidates before assembling a final page. While an item-level ranker estimates user response to each candidate, the reranking stage must select an ordered slate whose items interact. On mixed ads--organic (A--O) pages, the exposure of an advertisement may alter subsequent organic engagement, and the value of a given item may change with its position, nearby content, and earlier placements. The feasible slate is also constrained by ad load, spacing, eligibility, and protected positions. Consequently, sorting independently predicted item scores does not generally optimize whole-page value.

Existing reranking methods offer different compromises between contextual modeling and inference efficiency. Listwise networks represent interactions among candidates~\cite{ai2018dlcm,pei2019prm,pang2020setrank}, but predictions tied to the input arrangement need not generalize to an alternative displayed order. Non-autoregressive (NAR) allocation models amortize candidate--position scoring, but prefix-independent placement scores cannot directly represent how a candidate's value changes after a particular ordered sequence of exposures. NAR4Rec explicitly identifies missing target-item dependencies and introduces contrastive decoding as one remedy~\cite{ren2024nar4rec}; NGA incorporates global externalities but does not use the realized ordered prefix as an explicit input to every placement score~\cite{zheng2025nga}. Autoregressive generators capture permutation-level effects by conditioning on earlier decisions~\cite{bello2019seq2slate,zhu2024cga}; however, evaluating many constrained alternatives can be resource intensive when each candidate extension invokes a sequence scorer. These observations motivate a reranker that combines candidate-set context, ordered exposure history, and explicit slot context without repeatedly encoding the candidate pool.

We introduce Multi-Item Scoring (\method), a prefix-aware NAR-style scoring and constrained-generation framework that targets this specific gap between efficient, prefix-independent NAR allocation and sequence-conditioned AR generation. Its central idea is to compute shared candidate and slot representations once per request and then evaluate the marginal value of placing each remaining item after a selected prefix. Global self-attention captures interactions among all available candidates without assigning positional embeddings to their arbitrary input order. Localized prefix modeling captures exposure dependencies in the output order, from a direct predecessor to a full ordered history. Item--position cross-attention represents the slot where a candidate may be placed. The resulting scorer supports both supervised response prediction and constraint-aware slate search.

Long histories introduce a separate storage problem: retaining exact catalog
states scales with the number of item identities, while explicit score banks
also grow with the number of predecessor combinations. MIS addresses these
resource uses independently. Residual-quantized variational autoencoding (RQ-VAE)
represents historical states using a small shared vector bank and per-item
code tuples; active-row evaluation avoids enumerating the score tensor.
The former trades representation fidelity for storage, whereas the latter
preserves a fixed scorer's beam-search decisions. Current candidates remain
at full resolution.

A constraint-aware beam decoder reuses these request-level states and
evaluates only live prefix--candidate rows. Importantly, ``one-pass'' refers to shared candidate and slot encoding, not to one-step slate generation: prefix updates, less intensive conditional scoring, and beam selection remain sequential. Unlike a fully prefix-independent NAR scorer, MIS explicitly updates scores as the selected prefix changes; unlike a conventional AR decoder, it does not rerun intensive candidate-set encoding for each extension.

To test whether prefix conditioning closes the NAR--AR gap, we evaluate ranking prediction, evaluator-scored slate generation, prefix-length trade-offs, and decoding time on KuaiRand-1K and RL4RS. We then study five response-prediction tasks and four constrained A--O ordering regimes on industrial data, complemented by randomized evaluation readouts on a large-scale social-video platform. We distinguish offline model-scored improvements from observed online outcomes throughout. Recent industrial sequential ranking systems likewise underscore the importance of production-compatible shared-context computation and serving efficiency~\cite{hertel2026linkedin}.

Our main contributions are:
\begin{itemize}
\item We identify and address a central limitation of prefix-independent NAR whole-page allocation: its inability to directly condition placement values on the realized ordered prefix. MIS combines NAR-style shared candidate encoding with explicit local-prefix conditioning to narrow the quality gap to AR generation.
\item We develop ordered prefix conditioning with compact RQ-VAE historical descriptors and active-row constrained beam search. Shared request-level encoding is performed once, while lightweight prefix-dependent scores are recomputed for live beams; the design separately reduces catalog-state storage and avoids a dense combinatorial score bank.
\item We examine the accuracy--efficiency trade-off on two public datasets and evaluate multi-task prediction, constrained ads--organic ranking, and online system outcomes on a large-scale social platform.
\end{itemize}

\begin{figure*}[t]
  \centering
  \includegraphics[width=\textwidth]{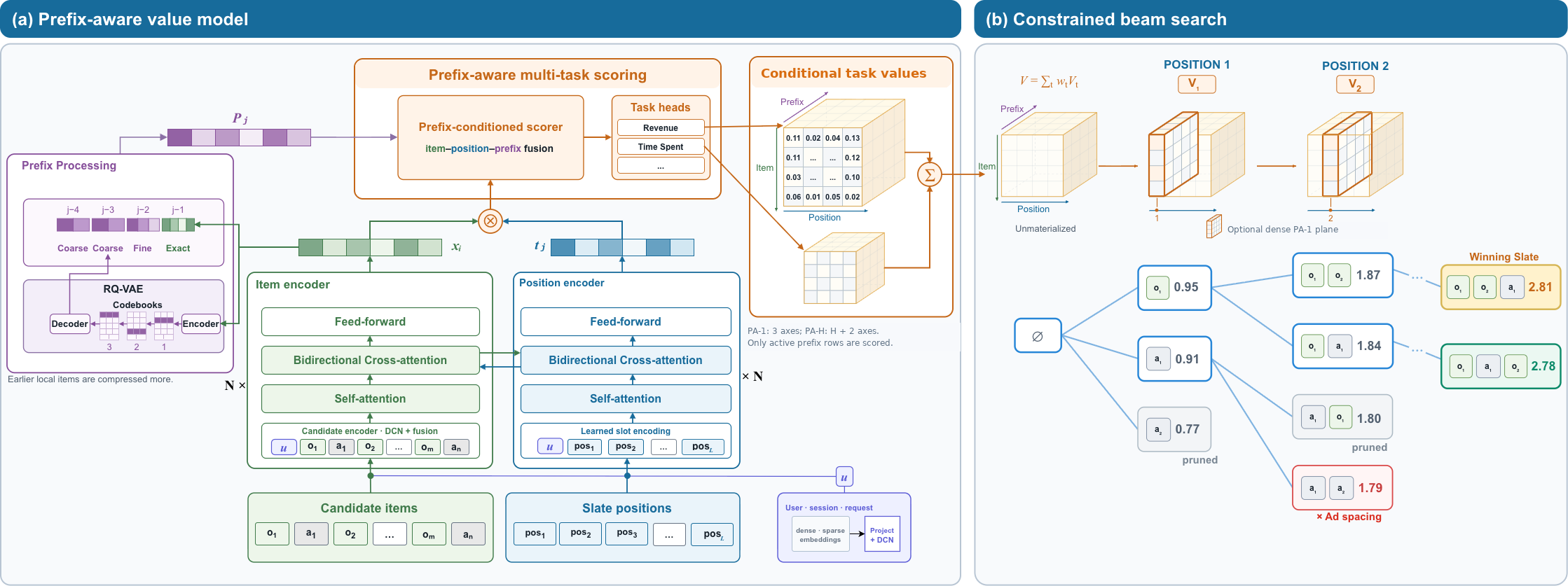}
  \caption{One-pass MIS and constrained generation. (a) Item and slot encoders
  supply global and position context to prefix-conditioned task heads; optional
  RQ-VAE descriptors preserve separate ordered lags. PA-$H$ has $H+2$ input
  axes; the cubes depict the optional PA-1 domain. (b) Task values guide beam
  search over active prefixes, with feasibility masks and top-$B$ selection.
  Only the $B\times C$ live score rows are materialized. Encoder repetition
  labels denote stack depth; numerical beam values are illustrative.}

  \Description{Shared attention-based item and position encoders feed
  prefix-conditioned task heads. The decoder combines task values and performs
  constrained beam search using only active-prefix score rows.}
  \label{fig:architecture}
\end{figure*}

\section{Related Work}
\paragraph{Context-aware reranking.}
DLCM uses recurrent list context~\cite{ai2018dlcm};
PRM and SetRank apply self-attention~\cite{pei2019prm,pang2020setrank}.
Other listwise and permutation-aware learning methods provide context for this design~\cite{cao2007listnet,xia2008listmle,carbonell1998mmr}.
These models evaluate a supplied list, whereas WPO must search feasible
orders whose values depend on prior placements.

\paragraph{Autoregressive generation.}
Seq2Slate generates permutations with a recurrent pointer
decoder~\cite{bello2019seq2slate}; CGA models permutation-level
externalities through autoregressive allocation~\cite{zhu2024cga}.
Autoregressive decoders can also share candidate representations.
Pointer networks and attention-based permutation models provide additional precedents for conditional sequence construction~\cite{vinyals2015pointer,vaswani2017attention}.
MIS instead exposes conditional item--slot values directly to constrained
search, separating request-level encoding from placement decisions.

\paragraph{Non-autoregressive models.}
NAR4Rec predicts item--position compatibility in parallel and explicitly notes the challenge of missing dependencies among target items; its contrastive decoding partially addresses this limitation~\cite{ren2024nar4rec}.
Other industrial and research systems explore sequential user modeling, contextual ranking, and off-policy recommendation~\cite{hertel2026linkedin,zhou2018din,zhou2019dien,chen2019topk}.
NGA also performs non-autoregressive allocation while modeling global
advertising externalities~\cite{zheng2025nga}.
The relevant distinction is not that all NAR models are necessarily less accurate than AR models, but that prefix-independent factorization cannot directly adapt each placement value to the realized ordered history. MIS retains NAR-style one-pass request encoding while explicitly conditioning placement values on the selected ordered prefix in mixed ads--organic WPO. CGA serves as a high-computation AR quality reference, whereas NAR/NGA are the primary efficiency-oriented comparisons.

\paragraph{Compact context.}
Residual quantization represents continuous states with hierarchical
codes~\cite{lee2022rqvae,jegou2011product,lloyd1982least}. Here codes compress historical descriptors;
they do not replace current candidates or generate catalog items.
Order remains explicit through separate predecessor coordinates.

\section{Problem Formulation}
For request context $c$, let $\mathcal{C}=\{1,\ldots,C\}$ be candidate items and
$L$ the output length. A slate is an ordered sequence
$\pi=(\pi_1,\ldots,\pi_L)$ with no repeated item. The feasible set
$\mathcal{F}(c)$ additionally enforces page constraints such as ad load,
minimum spacing, eligibility, and inventory-specific rules. We optimize
\begin{equation}
 \pi^*=\arg\max_{\pi\in\mathcal{F}(c)}
 \sum_{j=1}^{L} V(\pi_j,j,\pi_{<j};c).
 \label{eq:objective}
\end{equation}
A standard NAR model factorizes candidate and position states:
\begin{equation}
 \begin{aligned}
 (X,T)&=\operatorname{Enc}_{\theta}(c,\{q_i\}_{i=1}^C,\{1,\ldots,L\}),\\
 B(i,j)&=x_i^\top t_j,
 \end{aligned}
 \label{eq:base}
\end{equation}
where $x_i,t_j\in\mathbb{R}^{d}$ are the encoder states and $q_i$ contains
item and content-type features. The shared encoder may couple the two towers
through attention without conditioning on the generated prefix. Because $B(i,j)$ is
independent of $\pi_{<j}$, matching or beam search over $B$ cannot learn that
the same placement has different value after an ad versus an organic item.

The alternative is a full-list evaluator $E(c,\pi)$. It is expressive, but a
beam of width $B$ examines up to $B C$ continuations at each of $L$ steps,
requiring many candidate--prefix evaluations (possibly grouped into batched
invocations). Our goal is a conditional value that preserves the reuse in
Eq.~\ref{eq:base}.

\section{One-Pass MIS Model}
\label{sec:mis-model}
MIS separates reusable request encoding from order-dependent scoring.
Figure~\ref{fig:architecture} shows the item encoder, position encoder,
prefix processing, task heads, and constrained decoder.
The shared encoders produce $X=(x_1,\ldots,x_C)$ and
$T=(t_1,\ldots,t_L)$ once per request. Neither observed outcomes nor
future selections enter these representations. A conditional head combines
current-item, target-position, and prefix states:
\begin{equation}
 S_\tau(i,j,\pi_{<j};c)
 =B_\tau(i,j)+r_\tau(x_i,t_j,P_j).
 \label{eq:mis-score}
\end{equation}
Here $P_j$ represents the ordered context preceding position $j$;
Section~\ref{sec:context} specifies its construction.
The base $B_\tau(i,j)$ and candidate states are reused across all beams.
One-pass inference therefore means one shared candidate-encoder pass,
not a single constant-compute operation for the entire search.

\subsection{Multi-task ranking values}
A task head outputs a logit or regression score. Its link $\nu_\tau$
is sigmoid for binary outcomes and identity for continuous outcomes.
The page objective combines linked task values:
\begin{equation}
 V(i,j,\pi_{<j};c)=\sum_{\tau=1}^{Q}
 w_\tau(i,c)\,\nu_\tau(S_\tau(i,j,\pi_{<j};c)).
 \label{eq:multitask}
\end{equation}
Weights may incorporate fixed item-level coefficients.
This shared scoring interface supports two uses: evaluating proposed
slates and guiding their generation. Objective weights and feasibility rules remain fixed in
decoder comparisons.

\subsection{Training objective}
Given an observed slate $\pi$ and valid labels $y_{j,\tau}$, MIS learns
the score for the item, position, and preceding context actually observed:
\begin{equation}
 \mathcal{L}_{\mathrm{pred}}=\sum_{j,\tau}\omega_{j,\tau}
 \ell_\tau\!\left(y_{j,\tau},
 S_\tau(\pi_j,j,\pi_{<j};c)\right).
 \label{eq:training}
\end{equation}
For binary outcomes,
$\ell_{\mathrm{bin}}(y,s)=\log(1+\exp(s))-ys$; continuous targets use
$\ell_{\mathrm{reg}}(y,s)=(y-s)^2$.
Weights $\omega_{j,\tau}\geq0$ combine task weights and valid-label masks.
Training does not require labels for unobserved prefix combinations.
The prefix residual is initialized to zero, permitting a warm start from
the item--position base. Quantizer fitting uses only the permitted training
fold.

\section{Context-aware Modeling}
\label{sec:context}
Three complementary contexts determine a placement's value.
Global context captures interactions among all available candidates;
localized prefix context captures the order already selected; and position
context represents the slot being scored. Keeping these roles separate
prevents the initial candidate order from being mistaken for the output order.

\subsection{Global Context}
The global-context encoder jointly represents the complete candidate set and the request, allowing each candidate to attend to competing and complementary items before the output order is determined.
Self-attention permits an ad or organic item to represent its relationships
with the other available items:
\begin{equation}
 A_i=F_{\mathrm{item}}(q_i,c),\qquad
 X^{g}=G_{\mathrm{item}}(A_1,\ldots,A_C).
 \label{eq:global-context}
\end{equation}
No positional embedding is assigned to the input candidate indices: these indices specify an unordered pool, not the final ranking. This allows global attention to model cross-item interactions without assuming that the upstream retrieval order is the desired display order. Content type and
pre-allocation item features remain available.
With shared token-wise layers and unmasked set self-attention,
$G_{\mathrm{item}}(PA)=P G_{\mathrm{item}}(A)$ for a permutation matrix $P$.
Thus reindexing candidates reindexes their outputs rather than changing
their global interpretation. Output-position features enter the separate
position tower, not the initial list order.
Prefix conditioning separately identifies already placed items.

\subsection{Localized Context}
\label{sec:prefix}
We denote prefix-aware lookback-$H$ scoring by PA-$H$.
At position $j$, a lookback $H$ retains the ordered tuple
$\boldsymbol{\kappa}_j=(\kappa_{j,1},\ldots,\kappa_{j,H})$,
where $\kappa_{j,h}=\pi_{j-h}$ for $h<j$ and BOS otherwise.
Lag 1 is the immediate predecessor. BOS is a learned state with no observed
outcome. Local context models exposure fatigue, ad--organic substitution,
and other dependencies specific to the generated order.

\subsubsection{Short local context: reusing the immediate predecessor}
For $H=1$, the reusable predecessor projection $g_k=W_e x_k$, $k=\pi_{j-1}$,
directly conditions the current item; a learned $g_{\mathrm{BOS}}$ handles $j=1$:
\begin{equation}
 S^{(1)}(i,j,k)=B(i,j)+r_\psi([x_i;t_j;g_k]).
 \label{eq:tensor}
\end{equation}
This short-context variant directly reuses the predecessor representation, without a separate sequence encoder or quantization step. It uses the same beam
decoder as longer-prefix variants. A trilinear residual can be written as
$\sum_{a=1}^{r}(u_a^\top x_i)(v_a^\top t_j)(w_a^\top g_k)$;
an MLP instead maps the concatenated states nonlinearly.

\subsubsection{Longer prefixes preserve ordered coordinates}
If lag $h$ has $K_h$ descriptors including BOS, the conceptual score domain is
\begin{equation}
 \mathcal{V}^{(H)}_\tau\in
 \mathbb{R}^{C\times L\times K_1\times\cdots\times K_H}.
 \label{eq:high-order}
\end{equation}
There are $H+2$ axes: current item, target position, and $H$ independently
indexed predecessor coordinates. PA-1 is three-dimensional and PA-2
four-dimensional. Exact identities give $K_h=C+1$.
Complete-prefix PA-$N$ uses $N=L-1$ predecessor coordinates, with BOS
padding for shorter prefixes. A fixed-width representation of the tuple
does not reduce this high-order input domain to a three-dimensional tensor.
Only admissible active tuples are evaluated; the Cartesian product is
never enumerated.

A causal Transformer encodes the selected sequence without the current item:
\begin{equation}
 H_j=T_{\mathrm{causal}}([(F_e(x_{\pi_1})+e_1)^\top;\ldots;
 (F_e(x_{\pi_{j-1}})+e_{j-1})^\top]).
 \label{eq:pa-n}
\end{equation}
Prefix positional embeddings $e_\ell$ preserve the order actually selected;
they are distinct from the absent positional encoding of the initial
candidate list. A last-state head uses $H_{j,\mathrm{last}}$.
The reported candidate-conditioned head queries every ordered prefix state:
\begin{equation}
 \begin{aligned}
 a_{i,j}&=\operatorname{softmax}
 \left(\frac{q_i^{\mathrm{attn}}H_j^\top}{\sqrt d}\right)H_j,\\
 S_\tau^{(N)}&=B_\tau(i,j)+
 r_\tau([x_i;t_j;a_{i,j}^\top;x_i\odot a_{i,j}^\top]).
 \end{aligned}
 \label{eq:pa-n-attn}
\end{equation}
For $j>1$, $H_j\in\mathbb{R}^{(j-1)\times d}$,
$q_i^{\mathrm{attn}}=x_i^\top W_Q\in\mathbb{R}^{1\times d}$, and
$W_Q\in\mathbb{R}^{d\times d}$; the encoded states serve directly as
keys and values. $F_e$ is a learned affine prefix projection.
The $4d$-input residual MLP has hidden widths $2d$ and $d$.
For $j=1$, this implementation sets $a_{i,1}=0$ instead of running an
empty sequence. Prefix states are shared across candidates extending the
same beam; the item tower is not rerun.

\subsubsection{Efficient encoding of long prefixes}
Long-prefix scoring benefits from retaining item-specific history, but an
exact historical catalog of $U$ items and width $d$ requires $O(Ud)$ scalars.
Repeated identity coordinates also create a large potential score-index
space. We use RQ-VAE to replace the historical vector bank with shared
codebooks and compact assignments, rather than dropping history or averaging
away its order. Active-row decoding addresses the separate combinatorial
score storage. Quantization is useful primarily for large catalogs;
it is optional when exact history is affordable.
Historical-state storage is reduced by mapping selected lag coordinates to
semantic descriptors, without merging the lags.
For each selected predecessor, a quantizer maps its scorer state to an
ordered residual-code tuple. With $r$ retained levels, reconstruction is
$\tilde x_i^{(r)}=D_\xi(\sum_{s=1}^{r}q_{c_i^{(s)}}^{(s)})$.
The public configuration uses three levels with 16 primitive 32D vectors
per level: 48 stored vectors yield at most $16^3=4{,}096$ distinct
three-level tuples, plus BOS. Reducing reconstruction depth coarsens an
individual lag; it does not pool or permute the prefix.
For $R$ levels and $K_{\mathrm{prim}}$ total primitive vectors, the
historical catalog uses $O(K_{\mathrm{prim}}d+UR)$ storage, including code
assignments, rather than $O(Ud)$. The exact-history reference is Identity.
Table~\ref{tab:rq-storage} evaluates this storage--quality trade-off under
the same scorer; Appendix~\ref{app:quantizer-details} gives assignments
and the fitting loss.

The nearest predecessor may remain exact while older lags use fewer
reconstruction levels. Let $z_{j,h}$ denote lag $h$'s retained exact or
reconstructed state. The score is a joint function of the ordered sequence,
using a sequence encoder or explicit lag concatenation:
\begin{equation}
 S_\tau^{(H)}=B_\tau(i,j)+
 r_\tau([x_i;t_j;\mathcal{R}_H(z_{j,H},\ldots,z_{j,1})]).
 \label{eq:long-prefix}
\end{equation}
Quantization reduces descriptor cardinalities and storage while retaining
separate lag coordinates; decoding evaluates only active prefix tuples.
The representation ablation preserves its reported common reconstruction
protocol; the lag-wise resolution schedule illustrated in Figure~\ref{fig:architecture}
is configurable rather than imposed on every experimental variant.

\paragraph{Choosing the representation budget.}
Select history length and code resolution against both prediction and page
value, not reconstruction alone. Current candidate states remain exact.

\begin{table*}[t]
\centering
\caption{Five-seed prediction (12 epochs) and generation. AUC/NE use held-out
labels; a disjoint scorer evaluates beam-32 lift relative to random order.
Bold marks the best quality point estimate; underlining marks the runner-up.
$^\dagger$Separately fitted tower; Seq2Slate also fits its own evaluator.
Compute costs are full decoding by one seeded model at $B=32$, $L=C$:
profiled FLOPs (millions) and mean CPU latency (ms), as KuaiRand/RL4RS pairs.
Dashes denote prediction-only baselines. Appendix~\ref{app:implementation}
details architectures and aggregation. KuaiRand PA-$N$ uses Identity history
and is reproduced verbatim in Table~\ref{tab:rq-storage}; the PA-$N$ Identity reference in Appendix Table~\ref{tab:quantizers} also uses these figures.}
\label{tab:matched-results}
\small
\setlength{\tabcolsep}{3pt}
\begin{tabular}{lrrrrrrrr}
\toprule
& \multicolumn{3}{c}{KuaiRand-1K} & \multicolumn{3}{c}{RL4RS} & \multicolumn{2}{c}{Efficiency (K/R)} \\
\cmidrule(lr){2-4}\cmidrule(lr){5-7}\cmidrule(lr){8-9}
Model & AUC & NE & Lift & AUC & NE & Lift & FLOPs (M) & Latency (ms) \\
\midrule
DNN$^\dagger$ & 0.5597 & 0.9858 & -- & 0.6169 & 0.9324 & -- & -- & -- \\
DCN~\cite{wang2017dcn} & 0.6042 & 0.9757 & -- & 0.7696 & 0.8091 & -- & -- & -- \\
Seq2Slate$^\dagger$~\cite{bello2019seq2slate} & 0.6198 & 0.9673 & 16.325\% & 0.7865 & 0.7979 & 24.327\% & 14.86/14.37 & 1.00/0.91 \\
NAR~\cite{zheng2025nga} & 0.6038 & 0.9760 & 15.847\% & 0.7698 & 0.8081 & 23.697\% & 13.10/12.08 & 0.87/0.63 \\

CGA--AR ~\cite{zhu2024cga} & \textbf{0.6424} & \textbf{0.9553} & \underline{18.279\%} & \underline{0.8115} & \underline{0.7503} & \underline{33.276\%} & 650.50/476.79 & 27.54/20.28 \\
MIS--PA-1 & 0.6184 & 0.9693 & 16.954\% & 0.7804 & 0.7939 & 27.048\% & 19.99/16.20 & 1.26/1.08 \\
\textbf{MIS--PA-$N$} & \underline{0.6423} & \underline{0.9554} & \textbf{18.284\%} & \textbf{0.8123} & \textbf{0.7469} & \textbf{34.067\%} & 126.58/101.58 & 6.76/5.72 \\
\bottomrule
\end{tabular}
\end{table*}
\subsection{Position-aware candidate scoring}
\label{sec:position-context}
A separate position tower encodes each eligible output slot from its
learned slot embedding and request context:
$T^g=G_{\mathrm{pos}}(\{e_{\mathrm{slot}}(j)+F_{\mathrm{pos}}(c)\}_{j=1}^{L})$.
Item and position states interact through bidirectional cross-attention.
For one attention head, define
\begin{equation}
 \operatorname{CA}(Q,K,Z)=
 \operatorname{softmax}
 \left(\frac{(QW_Q)(KW_K)^\top}{\sqrt d}\right)(ZW_V).
 \label{eq:cross-attention}
\end{equation}
Then $\bar X=X^g+\operatorname{CA}(X^g,T^g,T^g)$ and
$\bar T=T^g+\operatorname{CA}(T^g,X^g,X^g)$, followed by
feed-forward blocks and the remaining encoder layers.
Their final states $x_i,t_j$ enter the base and prefix-conditioned heads.
These request-level interactions introduce slot context without initial
candidate ranks. They run once; only selected prefixes change in search.
\subsection{Holistic multi-task modeling}
\label{sec:holistic-mtml}
Whole-page optimization jointly considers several responses rather than a single click or conversion objective. We therefore use multi-task, multi-label learning (MTML) to share request-, candidate-, and position-level representations across task-specific prediction heads. Each head estimates its own response quantity given the same selected-prefix context, while the page optimizer combines linked predictions through Eq.~\ref{eq:multitask}. This separation allows ranking constraints and economic trade-offs to be specified at allocation time instead of being embedded implicitly in one prediction label.

For mixed ads--organic pages, heads predict ad impressions/clicks and
organic viewing/engagement. Fixed value coefficients enter the allocation
objective, not a learned label.
Section~\ref{sec:industrial} defines the industrial tasks.

\section{Constraint-Aware Slate Generation with One-Pass Inference}
\label{sec:active-prefix-search}
The intensive request-level item and position encoders run once. Search
then alternates ordered-prefix updates, conditional scoring, feasibility masking,
and beam selection. At position $j$, each of the $B$ live beams supplies one ordered prefix tuple.
The decoder fixes $j$ and that tuple, varies candidate $i$, and obtains a
length-$C$ score row. These rows form
$\widetilde V_j\in\mathbb{R}^{B\times C}$: selected entries of the
high-order tensor. The decoder materializes only this active-beam workspace,
rather than the full prefix tensor. Appendix~\ref{app:tensor-indexing}
illustrates the indexing.
Masks enforce uniqueness, eligibility, ad-load limits, HP-placement rules,
and minimum/maximum gaps before a bounded min-heap retains the best $B$
extensions. Score rows are discarded after selection. Beam-specific ordered descriptors, prefix states, and backpointers are
retained only for live beams. Every branch uses the same item and position
representations from the request-level encoder pass; these representations
do not depend on the order selected by search.

For a fixed trained model, active-row scoring gives the same extension values
as explicit tensor lookup:
\begin{equation}
 \widetilde V_j[b,i]
 = V(i,j,\pi_{<j}^{(b)};c).
 \label{eq:active-row}
\end{equation}
With identical feasibility masks, cumulative scores, and tie-breaking,
active-row evaluation preserves the fixed scorer's beam-search decisions.
Appendix~\ref{app:search-accounting} gives the equivalence argument and
the additional state requirements of pooled-tail variants.

Dense storage requires $O(LC\prod_{h=1}^{H}K_h)$ task values, or
$O(LC(C+1)^H)$ for exact histories. Active-row decoding instead uses
$O((C+L)d+LC+BC+BHd)$ persistent score/state storage: shared item and
position states, the base bank, live score rows, and ordered prefix states.
These bounds exclude weights and temporary encoder activations, with
task count and depth fixed. PA-$N$ uses $H=N$ and recomputes the
prefix attention states per live beam without rerunning the item tower.
Appendix~\ref{app:search-accounting} details alternative storage paths.

There are at most $BLC$ extensions. Masks and bounded-heap selection compute
$O(BLC\log(1+B))$ in the worst case; batched top-$B$ may use a different
selection algorithm. Let $F_I$ include shared encoding and base-bank
construction, $F_P(H)$ be a prefix-state
update, and $F_H(H)$ one conditional head evaluation. PA-$H$ requires
$F_I+O(BLF_P(H)+BLCF_H(H))$ neural work in addition to search.
One-pass refers to candidate encoding, not constant-time decoding or
the absence of prefix/head computation.

MIS supports four A--O regimes: AUOU preserves both relative orders;
ACOU reorders ads; AUOC reorders organics; and ACOC reorders both.

Compared with MIS, a prefix-independent NAR model uses $F_I$ shared neural
work and $O((C+L)d+LC+BC)$ persistent score/state storage. A candidate-wise autoregressive evaluator
requires $O(BLCF_S(H))$ neural work, where $F_S(H)$ is one candidate--prefix
sequence evaluation. All these decoders still incur constraint masking
and beam selection; shared request encoding removes repeated item-tower work, not the
combinatorial search problem.

\section{Experimental Setup}
\begin{figure*}[t]
\centering
\includegraphics[width=0.98\textwidth]{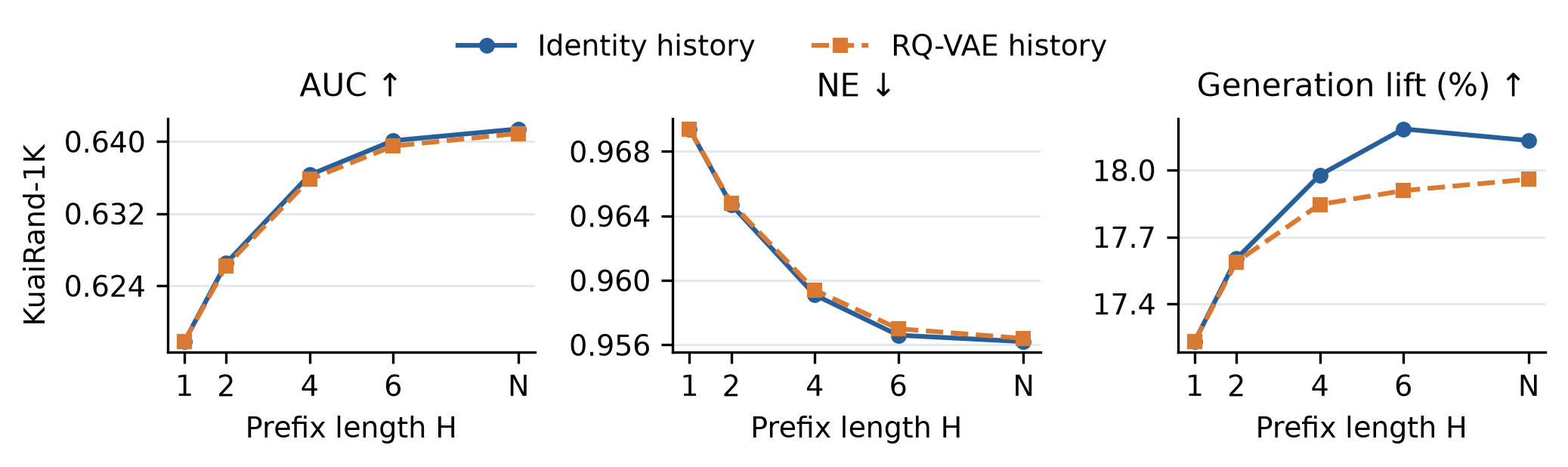}
\caption{RQ-VAE versus Identity history on KuaiRand-1K across prefix lengths.
The fixed length-sweep scorer uses exact historical vectors (Identity) or
RQ-VAE reconstructions. Panels report AUC, NE, and beam-32 generation lift
relative to random order. States coincide at $H=1$; $N=9$ denotes the full
prefix. This sweep is separate from the controlled full-prefix study in
Table~\ref{tab:rq-storage} and the candidate-query results in Table~\ref{tab:matched-results}.}
\label{fig:prefix-length}
\end{figure*}
\subsection{Datasets}
\paragraph{KuaiRand-1K}
KuaiRand contains randomly exposed short-video interactions and rich user/item
features~\cite{gao2022kuairand}. We construct 300,000 chronological windows of
ten impressions. Each position has one binary \texttt{is\_click} target; this is
a single-task, multi-position prediction problem. Retained video metadata form
a 68D input to the shared 32D item tower. We use an 80/20 pre-test/test split.
For counterfactual decoding, the pre-test portion is divided into disjoint
120,000-window generator and evaluator halves; 2,000 untouched test slates are
decoded.

\paragraph{RL4RS}
RL4RS Dataset A contains real nine-item slates with per-item feedback and item
attributes~\cite{wang2023rl4rs}. It provides a native-slate test of factual
prefix prediction and decoding. Each position has one binary feedback target;
the 40D item attributes feed the same 32D tower used by every method. We use the
first 300,000 records in released file order and preserve that order in the
80/20 split. The catalog contains only 284 item identities, complementing
KuaiRand's large-catalog regime.

\paragraph{Industrial AO whole-page data.}
The industrial dataset represents the A--O WPO setting in a large-scale short
video feed. Each request records the complete mixed page, including item order,
content type, position, and per-position ad impression/click and organic
view/engagement outcomes. A representative three-hour slice contains 36.7M
pages, with median length 20 and median five ads. Unlike homogeneous public
slates, these pages must jointly balance realization and user experience while
respecting eligibility, ad-load, HP-placement, and inter-ad gap rules. Prediction trains on seven days and evaluates the next day; decoding
uses 300K pages with fixed candidate pools and constraints.
Only aggregate normalized metrics are reported.

\begin{table}[t]
\centering
\caption{Matched KuaiRand PA-$N$ compression. The Identity baseline (AUC 0.6423, NE 0.9554, lift 18.284\%) exactly matches the KuaiRand MIS--PA-$N$ row of Table~\ref{tab:matched-results}; only historical representations change.
Catalog payload assumes FP32 vectors and packed 4-bit assignments,
excluding networks and working states. Storage scales as $O(Ud)$ for
Identity and $O(K_{\mathrm{prim}}d+UR)$ for RQ-VAE.}
\label{tab:rq-storage}
\small
\setlength{\tabcolsep}{2.5pt}
\begin{tabular}{lrrrr}
\toprule
History & AUC & NE & Lift (\%) & MiB\\
\midrule
Identity & 0.6423 & 0.9554 & 18.284 & 187.81\\
RQ-VAE & 0.6418 & 0.9557 & 18.120 & 2.207\\
\bottomrule
\end{tabular}
\end{table}
\subsection{Baselines and variants}
We compare DNN, DCN, Seq2Slate, NAR, CGA, and MIS. NAR is a prefix-independent item--position allocation
model inspired by NGA~\cite{zheng2025nga}; CGA~\cite{zhu2024cga} uses a causal
Transformer that re-encodes each candidate prefix; Seq2Slate~\cite{bello2019seq2slate}
uses a recurrent pointer generator trained on logged binary permutation targets.
DNN uses a one-dimensional trainable bottleneck over
a separately fitted frozen item tower; DCN~\cite{wang2017dcn} supplies explicit feature-cross
interactions. NAR and CGA compare
allocation architectures. We implement the baselines within the public-data
prediction and allocation framework.
NAR, DCN, CGA, and MIS share each study's frozen supervised 32D tower.
DNN and Seq2Slate use separately fitted towers with the same input
features and fold boundaries.
The public study isolates prefix scoring over frozen catalog states;
the industrial architecture additionally uses request-level global and slot
attention. Appendix~\ref{app:implementation} gives widths, losses, and aggregation.

MIS--PA-$N$ uses complete-prefix candidate-query attention
(Eq.~\ref{eq:pa-n-attn}). Table~\ref{tab:rq-storage} changes only historical
reconstructions under its identical five-seed protocol; Identity repeats
Table~\ref{tab:matched-results}. Supplementary sweeps use separate last-state
heads. Industrial comparisons use pointwise, NAR, MIS-PA1, and MIS-PA4,
with MIS-PA4 used for generation and the industrial scalability experiment.

The robustness study uses \emph{PA-4 + pooled tail}: exact lag 1,
three-level RQ lag 2, two-level RQ lags 3--4, and a running mean of
reconstructed complete-prefix states. Unlike PA-$N$, it cannot distinguish
permutations outside its four ordered coordinates.

\begin{table*}[t]
\centering
\caption{Industrial A--O results. MIS-PA4 is the candidate used for page
generation and the scalability experiment; MIS-PA1 is the first-order variant.
(a) All five displayed next-day prediction tasks use normalized entropy (NE).
Organic 15-second view and valuable view correspond to VPV and VPVD.
Changes are
relative to pointwise. (b) Beam-64 page-value lift (\%) over logged order under
identical constraints. Bold values mark the lowest prediction error or highest
lift, including ties.}
\label{tab:prod-pred}
\small
\setlength{\tabcolsep}{2pt}
\begin{minipage}[t]{0.66\textwidth}
\centering
\begin{tabular}{lrrrr}
\toprule
\multicolumn{5}{l}{\textit{(a) Next-day prediction}} \\
Task & Pointwise & NAR & MIS-PA1 & MIS-PA4 \\
\midrule
Impression & 0.7641 & 0.7623 ($-0.23\%$) & 0.7601 ($-0.52\%$) & \textbf{0.7597 ($-0.58\%$)} \\
Ad click & 0.7655 & 0.7648 ($-0.09\%$) & 0.7644 ($-0.14\%$) & \textbf{0.7642 ($-0.17\%$)} \\
Organic share & 0.4542 &  0.4512 ($-0.66\%$)  &0.4504 ($-0.84\%$) & \textbf{0.4503 ($-0.86\%$)} \\
Organic 15-second view & 0.7452  & 0.7439 ($-0.17\%$) & 0.7438 ($-0.19\%$)& \textbf{0.7436 ($-0.21\%$)} \\
Organic valuable view & 0.9486 & 0.9485 ($-0.01\%$) & 0.9484 ($-0.02\%$) & \textbf{0.9483 ($-0.03\%$)} \\
\bottomrule
\end{tabular}
\end{minipage}\hfill
\begin{minipage}[t]{0.32\textwidth}
\centering
\begin{tabular}{lrrrr}
\toprule
\multicolumn{5}{l}{\textit{(b) Page-value lift (\%)}} \\
Decoder & AUOU & ACOU & AUOC & ACOC \\
\midrule
NAR & 0.444 & 0.435 & 4.119 & 4.245 \\
MIS-PA4 & \textbf{0.605} & \textbf{0.632} & \textbf{4.424} & \textbf{4.462} \\
\bottomrule
\end{tabular}
\end{minipage}
\end{table*}
\begin{table}[t]
\centering
\caption{Industrial scale evaluation results show relative changes (\%) versus an industrial ranker; brackets
give 95\% confidence intervals. Outcome and seven-day serving metrics
use their specified populations (Appendix~\ref{app:implementation}).
Negative service deltas indicate lower CPU consumption.
These are service, not isolated model-inference, metrics.}
\label{tab:online}
\small
\begin{tabular}{lr}
\toprule
Metric & Result \\
\midrule
Realization metric & $+0.104$ [$+0.054$, $+0.153$] \\
Overall video watch time & $+0.058$ [$+0.001$, $+0.114$] \\
Linked sessions & $-0.002$ [$-0.054$, $+0.055$] \\
\midrule
\multicolumn{2}{l}{\emph{Serving metrics: relative change (\%)}} \\
Ranking-service CPU (mean) & $-9.73$ [$-10.06$, $-9.46$] \\
Slate-evaluation CPU & $-9.73$ [$-10.13$, $-9.40$] \\
Prediction-service CPU (mean) & $+1.39$ [$+0.30$, $+2.54$] \\
\bottomrule
\end{tabular}
\end{table}
\subsection{Evaluation protocol}
Prediction uses pooled AUC and normalized entropy (NE; lower is better).
Headline results use five seeds and 12 epochs; supplementary representation
and robustness studies retain three seeds.

For a binary task with held-out predictions $p_n$ and labels $y_n$, we use
\begin{equation}
 \mathrm{NE}=\frac{-M^{-1}\sum_{n=1}^{M}
 [y_n\log p_n+(1-y_n)\log(1-p_n)]}
 {-\bar y\log\bar y-(1-\bar y)\log(1-\bar y)},
 \label{eq:ne}
\end{equation}
where $\bar y=M^{-1}\sum_n y_n$ and $0<\bar y<1$. All variants share
the same evaluation examples and task-specific denominator. AUC pools the
same examples; its ranking criterion need not agree exactly with NE.
Reordered slates are evaluated by a disjoint five-model causal Transformer
ensemble; robustness studies retain three models. Reversing
generator/evaluator folds and substituting a causal GRU test fold and
architecture sensitivity. Intervals use 2,000 paired slate bootstrap
resamples. Independent fitting reduces leakage but does not make
model-scored outcomes causal effects
\cite{schnabel2016recommendations,joachims2017counterfactual}.

On held-out labels, the headline evaluator obtains
AUC $0.6192$ and NE $0.9697$ on KuaiRand, versus AUC $0.8363$ and
NE $0.7417$ on RL4RS. The evaluator tower fits only its own half;
neither states nor labels are shared with generators.

Decoder profiling uses one AMD EPYC-Genoa thread, ten warmups, and
100 timed calls at beam width 32. It includes conditional scoring,
masking, and beam selection with seeded weights; Appendix~\ref{app:search-accounting}
specifies the length sweep and FLOP-accounting protocol.

\section{Public-Data Ranking Results}

\subsection{Prediction and Generation Quality}
Table~\ref{tab:matched-results} reports held-out prediction and
model-scored generation under the beam budgets stated in its caption.
MIS--PA-$N$ improves all three metrics over NAR and PA-1.
Against CGA on KuaiRand, its AUC is 0.0001 lower and NE is 0.0001 higher,
while generation lift is $0.005\pp$ higher. On RL4RS, AUC is 0.0008 higher
and NE is 0.0034 lower, while generation lift is $0.791\pp$ higher.
MIS--PA-$N$ therefore achieves comparable prediction quality and slightly higher generation lift than CGA on KuaiRand, and outperforms it on all three RL4RS quality point estimates.
At the same beam width, the single-model decoding benchmark reports
$4.07\times$ and $3.54\times$ lower latency on KuaiRand and RL4RS,
respectively (Section~\ref{sec:efficiency}), yielding a favorable
quality--efficiency trade-off. This comparison does not establish
statistical equivalence between the methods.

CGA jointly re-encodes each candidate with its prefix, whereas MIS shares
candidate representations and applies prefix-conditioned scoring separately.
This different factorization is a possible interpretation of the observed
trade-off, not a demonstrated cause of the quality gaps.
Table~\ref{tab:matched-results} uses Identity history for PA-$N$; the
separate RQ-VAE trade-off is evaluated in Table~\ref{tab:rq-storage}.
Appendix~\ref{app:implementation} specifies seed aggregation.

\subsection{Search Robustness}
A separate KuaiRand beam-32 study compares NAR with the PA-4 + pooled-tail variant
defined in the setup. Their primary-evaluator lifts are 15.931\% and 17.938\%:
a paired difference of $2.007\pp$ (95\% CI $[1.842,2.167]$).
Reversing generator/evaluator folds yields 13.208\% versus 14.600\%,
a paired gap of $1.393\pp$ (CI $[1.260,1.521]$).
A causal-GRU evaluator yields 16.162\% versus 18.578\%,
a gap of $2.416\pp$ (CI $[2.235,2.594]$).
Differences are computed before rounding.

NAR decreases from 18.09\% lift at beam 1 to 15.93\% at beam 32, whereas
PA-4 + pooled tail remains nearly stable at 18.12--17.94\%. The consistent direction under altered folds and evaluators supports
search alignment across these KuaiRand protocols.

\subsection{RQ-VAE Compression}
\label{sec:prefix-length}
\label{sec:quantizers}

Table~\ref{tab:rq-storage} compares RQ-VAE directly with Identity under the
same candidate-query PA-$N$ configuration used in Table~\ref{tab:matched-results}.
On KuaiRand, RQ-VAE changes AUC by $-0.0005$, NE by
$+0.0003$, and generation lift by $-0.164\pp$: close point estimates,
rather than a demonstrated equivalence test. Identity retains 1.54M exact
32D vectors; RQ-VAE uses 48 primitive vectors and three code indices per item.
The vector bank is about $32{,}053\times$ smaller. Including analytically
packed assignments, the represented catalog payload decreases from
187.81 to 2.207 MiB, about $85.1\times$; wider code-index storage reduces
this ratio. These are historical-catalog payloads, not end-to-end memory
savings or reductions in the number of independent prefix axes.

Figure~\ref{fig:prefix-length} complements the fixed full-prefix comparison.
As exact lookback increases from one predecessor to complete history,
KuaiRand AUC rises from 0.6178 to 0.6414 and NE falls from 0.9694 to 0.9562.
RQ-VAE tracks this trajectory: the maximum AUC and NE gaps are 0.0006 and
0.0004, respectively, in this separate sweep. Generation lift is not monotonic
and full-prefix RQ-VAE lift is $0.174\pp$ below Identity.
The length sweep uses its own last-state scorer and is not pooled with
Tables~\ref{tab:matched-results}--\ref{tab:rq-storage}; the latter share
one matched candidate-query PA-$N$ result.
Thus compact descriptors retain much of the large-catalog history signal
without requiring an exact vector per catalog item.
Appendix Table~\ref{tab:quantizers} presents the matched RQ-VAE and Identity results alongside a K-means result from a separate fixed last-state scoring protocol.

\subsection{Efficiency}
\label{sec:efficiency}
Table~\ref{tab:matched-results} reports measured full-decoding compute.
At $B=32$, single-model PA-$N$ takes 6.763/5.721 ms on KuaiRand/RL4RS
versus CGA's 27.543/20.280 ms: $4.07\times$/$3.54\times$ speedups.
The corresponding arithmetic counts are 126.58/101.58 versus
650.50/476.79 million FLOPs. NAR, PA-1, and Seq2Slate provide cheaper
decoding paths with the quality shown in the same table.

MIS reuses candidate states and shares each live prefix encoding across
extensions; CGA re-encodes candidate--prefix sequences. PA-$N$ still
performs conditional scoring and search at each step. Active rows require
$O(LC+BC)$ scores plus live prefix states, rather than a dense
$CL\prod_hK_h$ bank. Scalability additionally depends on feature
access, batching, constraints, and candidate count.
Appendix~\ref{app:implementation} gives timing intervals, ensembles,
and operation-accounting details.

\section{Industrial Ads--Organic Whole-Page Reranking}
\label{sec:industrial}
\subsection{Prediction}
The industrial multi-task ranker predicts ad impressions/clicks and organic
share, 15-second view (VPV), and valuable view (VPVD).
VPV is the binary 15-second-view target; VPVD is the platform-defined binary
valuable-view target. Definitions remain fixed across variants.

Table~\ref{tab:prod-pred}(a) reports the five displayed held-out
next-day prediction tasks. Parenthesized changes use the pointwise value as the
denominator; lower errors are better. Panel (b) evaluates constrained page
generation with the same MIS-PA4 model.

MIS has lower NE point estimates for all five displayed tasks, including
impression NE of 0.7597 versus pointwise's 0.7641.
Ad-click and organic valuable-view gains are modest; without uncertainty intervals,
the table does not establish significance for every task.

\subsection{Constrained ads--organic reranking}
We combine predicted outcomes into a holistic page-ranking objective. For an ad at position $j$,
the score includes an estimated realization component based on applicable item-level value coefficients and impression terms; for organic content,
it includes valuable-view and 15-second-view terms. All methods use identical
weights and candidate pools. Panel (b) of Table~\ref{tab:prod-pred} reports
beam-64 generation alongside the prediction baselines. Prefix awareness
improves predicted page value in every regime by $0.16$--$0.31\pp$ beyond the
NAR decoder.

In the fully reorderable ACOC regime, the model-scored decomposition estimates
a $+22.2\%$ improvement in realization score relative to the logged order, alongside 2.6\% lower ad
load, $+3.0\%$ 15-second views, and $+2.8\%$ valuable views. Relative to
NAR, the estimated realization-score difference is $+6.3\%$, with the reported ad-load
and engagement differences within 1\%. These are \emph{in-model estimates}
from the scorer used by the decoder; they indicate optimization headroom,
not independently observed outcomes or realized online lift.
The randomized study below provides a separate observed-outcome comparison.

The feasible search space is substantial even at moderate ad load: for a
20-item page, the evaluated spacing rules allow 4,275 AUOU layouts with five
ads, 2,554 with four ads, and 851 with three ads. For the 300K-page offline
study, beam-64 slate assembly takes about 4.5 minutes per regime on 70 CPU
cores. This batch throughput is distinct from per-request serving latency.

\subsection{Randomized industrial scale evaluation}
\label{sec:online}
The randomized study evaluates MIS on a large-scale social-video platform.
It compares the industrial-scale MIS candidate with an industrial ranker, using 300 hash-assigned segments per arm.
The launch rule requires a realization improvement without guardrail
regression. Table~\ref{tab:online} combines observed outcomes and a
seven-day serving readout; metric populations and assignment details are
specified in Appendix~\ref{app:implementation}.

Table~\ref{tab:online} shows positive realization and watch-time effects; the
linked-session interval includes zero. Ranking-service CPU falls $9.73\%$ (95\% confidence interval excluding zero), while prediction-service CPU rises $1.39\%$.
Slate-evaluation CPU is part of ranking-service CPU, not an additional saving.

The additional prefix-conditioned predictor computation may increase
prediction-service CPU, while one-pass shared representations and conditional
placement scores can avoid repeated ranking-stage evaluations. These are
plausible explanations for the opposite CPU trends, not mechanisms isolated
by the aggregate service measurements. The ranking-service reduction
($-9.73\%$) and prediction-service increase ($+1.39\%$) are measured
relative to their respective service baselines; they do not by themselves
establish a net CPU reduction across both services. Percentages have different
baseline resource volumes and cannot be summed into a total infrastructure
saving. They establish a service-level trade-off, not uniformly less intensive
inference. Objective weights and feasibility rules remain in the allocation layer.

\section{Conclusion}
MIS separates one-pass request encoding from conditional whole-page
scoring. Global candidate interactions, ordered prefix states, and
item--position context support constrained generation without re-encoding
the candidate pool. Active rows avoid materializing the high-order score
domain. Public experiments demonstrate complete-prefix prediction and
generation quality approaching adapted width-matched CGA, with
$4.07\times$/$3.54\times$ faster single-model decoding on KuaiRand/RL4RS.
The industrial study connects multi-task prediction and constrained A--O
allocation with observed online outcomes and serving-system measurements.
The interface exposes history resolution, search compute, and inference compute.

\label{sec:main-end}
\FloatBarrier

\bibliographystyle{ACM-Reference-Format}
\bibliography{references}

@inproceedings{wang2017dcn,
  title={Deep \& Cross Network for Ad Click Predictions},
  author={Wang, Ruoxi and Fu, Bin and Fu, Gang and Wang, Mingliang},
  booktitle={Proceedings of the ADKDD'17},
  year={2017},
  doi={10.1145/3124749.3124754}
}

@article{bello2019seq2slate,
  title={Seq2Slate: Re-ranking and Slate Optimization with RNNs},
  author={Bello, Irwan and Kulkarni, Sayali and Jain, Sagar and Boutilier, Craig and Chi, Ed and Eban, Elad and Luo, Xiyang and Mackey, Alan and Meshi, Ofer},
  journal={arXiv preprint arXiv:1810.02019},
  year={2019}
}

@inproceedings{ren2024nar4rec,
  title={Non-autoregressive Generative Models for Reranking Recommendation},
  author={Ren, Yuxin and Yang, Qiya and Wu, Yichun and Xu, Wei and Wang, Yalong and Zhang, Zhiqiang},
  booktitle={Proceedings of the 30th ACM SIGKDD Conference on Knowledge Discovery and Data Mining},
  year={2024}
}

@article{zheng2025nga,
  title={NGA: Non-autoregressive Generative Auction with Global Externalities for Advertising Systems},
  author={Zheng, Zuowu and Wang, Ze and Yang, Fan and Ye, Wenqing and Huang, Weihua and He, Wenqiang and Zhang, Teng and Wang, Xingxing},
  journal={arXiv preprint arXiv:2506.05685},
  year={2025}
}

@article{zhu2024cga,
  title={Contextual Generative Auction with Permutation-level Externalities for Online Advertising},
  author={Zhu, Ruitao and Liu, Yangsu and Chen, Dagui and Ma, Zhenjia and Shi, Chufeng and Zheng, Zhenzhe and Zhang, Jie and Xu, Jian and Zheng, Bo and Wu, Fan},
  journal={arXiv preprint arXiv:2412.11544},
  year={2024}
}

@inproceedings{ai2018dlcm,
  title={Learning a Deep Listwise Context Model for Ranking Refinement},
  author={Ai, Qingyao and Bi, Keping and Guo, Jiafeng and Croft, W. Bruce},
  booktitle={Proceedings of the 41st International ACM SIGIR Conference on Research and Development in Information Retrieval},
  year={2018},
  doi={10.1145/3209978.3209985}
}

@inproceedings{pei2019prm,
  title={Personalized Re-ranking for Recommendation},
  author={Pei, Changhua and Zhang, Yi and Zhang, Yongfeng and Sun, Fei and Lin, Xiao and Sun, Hanxiao and Wu, Jian and Jiang, Peng and Ou, Wenwu},
  booktitle={Proceedings of the 13th ACM Conference on Recommender Systems},
  year={2019}
}

@inproceedings{pang2020setrank,
  title={SetRank: Learning a Permutation-Invariant Ranking Model for Information Retrieval},
  author={Pang, Liang and Xu, Jun and Ai, Qingyao and Lan, Yanyan and Cheng, Xueqi and Wen, Jirong},
  booktitle={Proceedings of the 43rd International ACM SIGIR Conference on Research and Development in Information Retrieval},
  year={2020}
}

@inproceedings{wang2023rl4rs,
  title={RL4RS: A Real-World Dataset for Reinforcement Learning based Recommender System},
  author={Wang, Kai and Zou, Zhene and Zhao, Minghao and Deng, Qilin and Shang, Yue and Liang, Yile and Wu, Runze and Shen, Xudong and Lyu, Tangjie and Fan, Changjie},
  booktitle={Proceedings of the 46th International ACM SIGIR Conference on Research and Development in Information Retrieval},
  year={2023}
}

@inproceedings{gao2022kuairand,
  title={KuaiRand: An Unbiased Sequential Recommendation Dataset with Randomly Exposed Videos},
  author={Gao, Chongming and Li, Shijun and Zhang, Yuan and Chen, Jiawei and Li, Biao and Lei, Wenqiang and Jiang, Peng and He, Xiangnan},
  booktitle={Proceedings of the 31st ACM International Conference on Information and Knowledge Management},
  year={2022},
  doi={10.1145/3511808.3557624}
}

@inproceedings{vaswani2017attention,
  title={Attention Is All You Need},
  author={Vaswani, Ashish and Shazeer, Noam and Parmar, Niki and Uszkoreit, Jakob and Jones, Llion and Gomez, Aidan N. and Kaiser, Lukasz and Polosukhin, Illia},
  booktitle={Advances in Neural Information Processing Systems},
  year={2017}
}

@inproceedings{lee2022rqvae,
  title={Autoregressive Image Generation Using Residual Quantization},
  author={Lee, Doyup and Kim, Chiheon and Kim, Saehoon and Cho, Minsu and Han, Wook-Shin},
  booktitle={Proceedings of the IEEE/CVF Conference on Computer Vision and Pattern Recognition},
  year={2022}
}

@article{lloyd1982least,
  title={Least Squares Quantization in PCM},
  author={Lloyd, Stuart P.},
  journal={IEEE Transactions on Information Theory},
  volume={28},
  number={2},
  pages={129--137},
  year={1982}
}

@article{jegou2011product,
  title={Product Quantization for Nearest Neighbor Search},
  author={J{\'e}gou, Herv{\'e} and Douze, Matthijs and Schmid, Cordelia},
  journal={IEEE Transactions on Pattern Analysis and Machine Intelligence},
  volume={33},
  number={1},
  pages={117--128},
  year={2011}
}

@inproceedings{schnabel2016recommendations,
  title={Recommendations as Treatments: Debiasing Learning and Evaluation},
  author={Schnabel, Tobias and Swaminathan, Adith and Singh, Ashudeep and Chandak, Navin and Joachims, Thorsten},
  booktitle={Proceedings of the 33rd International Conference on Machine Learning},
  year={2016}
}

@inproceedings{joachims2017counterfactual,
  title={Unbiased Learning-to-Rank with Biased Feedback},
  author={Joachims, Thorsten and Swaminathan, Adith and Schnabel, Tobias},
  booktitle={Proceedings of the Tenth ACM International Conference on Web Search and Data Mining},
  year={2017}
}

@article{hertel2026linkedin,
 title={An Industrial-Scale Sequential Recommender for LinkedIn Feed Ranking},
 author={Hertel, Lars and Srivastava, Gaurav and Naqvi, Syed Ali and Kumar, Satyam and Zhang, Yue and others},
 journal={arXiv preprint arXiv:2602.12354},
 year={2026}
}

@inproceedings{cao2007listnet,
 title={Learning to Rank: From Pairwise Approach to Listwise Approach},
 author={Cao, Zhe and Qin, Tao and Liu, Tie-Yan and Tsai, Ming-Feng and Li, Hang},
 booktitle={Proceedings of the 24th International Conference on Machine Learning}, year={2007}, doi={10.1145/1273496.1273513}}

@inproceedings{xia2008listmle,
 title={Listwise Approach to Learning to Rank: Theory and Algorithm},
 author={Xia, Fen and Liu, Tie-Yan and Wang, Jue and Zhang, Wensheng and Li, Hang},
 booktitle={Proceedings of the 25th International Conference on Machine Learning}, year={2008}, doi={10.1145/1390156.1390306}}

@inproceedings{carbonell1998mmr,
 title={The Use of MMR, Diversity-Based Reranking for Reordering Documents and Producing Summaries},
 author={Carbonell, Jaime and Goldstein, Jade},
 booktitle={Proceedings of the 21st Annual International ACM SIGIR Conference}, year={1998}, doi={10.1145/290941.291025}}

@inproceedings{vinyals2015pointer,
 title={Pointer Networks}, author={Vinyals, Oriol and Fortunato, Meire and Jaitly, Navdeep},
 booktitle={Advances in Neural Information Processing Systems}, year={2015}}

@inproceedings{zhou2018din,
 title={Deep Interest Network for Click-Through Rate Prediction},
 author={Zhou, Guorui and Zhu, Xiaoqiang and Song, Chenru and Fan, Ying and Zhu, Han and Ma, Xiao and Yan, Yanghui and Jin, Junqi and Li, Han and Gai, Kun},
 booktitle={Proceedings of the 24th ACM SIGKDD International Conference on Knowledge Discovery and Data Mining}, year={2018}, doi={10.1145/3219819.3219823}}

@inproceedings{zhou2019dien,
 title={Deep Interest Evolution Network for Click-Through Rate Prediction},
 author={Zhou, Guorui and Mou, Na and Fan, Ying and Pi, Qi and Bian, Weijie and Zhou, Chang and Zhu, Xiaoqiang and Gai, Kun},
 booktitle={Proceedings of the AAAI Conference on Artificial Intelligence}, year={2019}, doi={10.1609/aaai.v33i01.33015944}}

@inproceedings{chen2019topk,
 title={Top-K Off-Policy Correction for a REINFORCE Recommender System},
 author={Chen, Minmin and Beutel, Alex and Covington, Paul and Jain, Sagar and Belletti, Francois and Chi, Ed H.},
 booktitle={Proceedings of the Twelfth ACM International Conference on Web Search and Data Mining}, year={2019}, doi={10.1145/3289600.3290999}}

\clearpage
\appendix
\setcounter{figure}{0}
\renewcommand{\thefigure}{A\arabic{figure}}
\renewcommand{\theHfigure}{appendix.\arabic{figure}}
\twocolumn[{
\begin{minipage}{\textwidth}
\section{Supplementary Prefix Representation Details}
\label{app:prefix-sweep}

\centering
\captionof{table}{KuaiRand-1K compact-history results. RQ-VAE and Identity reproduce the matched PA-$N$ evaluation from Table~\ref{tab:rq-storage}; K-means comes from a separate fixed last-state scorer and should not be compared directly with these PA-$N$ rows. Descriptor counts denote occupied codes (Identity: catalog capacity); The K-means row is descriptive only; comparisons against the matched PA-$N$ rows are not controlled.}
\label{tab:quantizers}
\small
\setlength{\tabcolsep}{2.5pt}
\begin{tabular}{llrrrrr}
\toprule
Dataset & Representation & Vector bank & Descriptors (pred/beam) & AUC & NE & Lift \\
\midrule
\multirow{3}{*}{KuaiRand-1K}
 & K-means & $48\times32$ & 48/48 & 0.6404 & 0.9570 & 17.977\% \\
 & RQ-VAE & $48\times32$ & 1,809/3,690 & 0.6418 & 0.9557 & 18.120\% \\
 & Identity (PA-$N$ ref.) & $1.54\mathrm{M}\times32$ & 1.54M/1.54M & 0.6423 & 0.9554 & 18.284\% \\
\bottomrule
\end{tabular}
\end{minipage}
\vspace{1em}
}]

Table~\ref{tab:quantizers} reports KuaiRand-only compact-history results.
The RQ-VAE row uses the same matched PA-$N$ evaluation as
Table~\ref{tab:rq-storage} (AUC 0.6418, NE 0.9557, lift 18.120\%),
and the Identity row uses the identical PA-$N$ reference (0.6423,
0.9554, and 18.284\%). Within this matched configuration, RQ-VAE
reduces AUC by 0.0005, increases NE by 0.0003, and decreases generation
lift by $0.164\pp$ relative to Identity; these are point estimates.
The K-means row comes from a separate fixed last-state scorer and is
included as a descriptive compact-code baseline only. Its AUC, NE, and
lift must not be used to infer a controlled RQ-VAE versus K-means
improvement. Descriptor counts summarize occupied historical codes in
prediction and beam evaluation; they do not change the scoring-protocol
qualification.

On KuaiRand, the separate prefix-length experiment in
Figure~\ref{fig:prefix-length} varies the historical lookback under exact
and RQ-VAE encoding. The full prefix contains nine predecessors;
length one uses the same direct state in both paths.
Compression changes AUC by at most 0.0006 and NE by at most 0.0004;
full-prefix RQ-VAE generation lift is $0.174\pp$ below exact history.
This length sweep and Table~\ref{tab:quantizers} use distinct protocols
from the matched PA-$N$ comparison, so their numerical results should
not be pooled.

Catalog identity storage is $O(Ud)$ for $U$ items. A codebook with
$K_{\mathrm{prim}}=48$ primitive vectors and $R=3$ levels uses
$O(K_{\mathrm{prim}}d+UR)$ storage, including assignments but excluding
quantizer parameters. On KuaiRand, the catalog vector-count ratio is
approximately $32{,}053\times$.
This is not an end-to-end memory reduction: current candidate states,
code assignments, and network parameters remain necessary.
These catalog comparisons use fixed, context-independent tower states.
A request-conditioned encoder instead requires descriptors consistent
with the current request; item identity alone cannot retrieve its
contextual state.

\clearpage
\twocolumn[{
\begin{minipage}{\textwidth}
\section{Active-row Tensor Indexing}
\label{app:tensor-indexing}
\centering
\includegraphics[width=\textwidth]{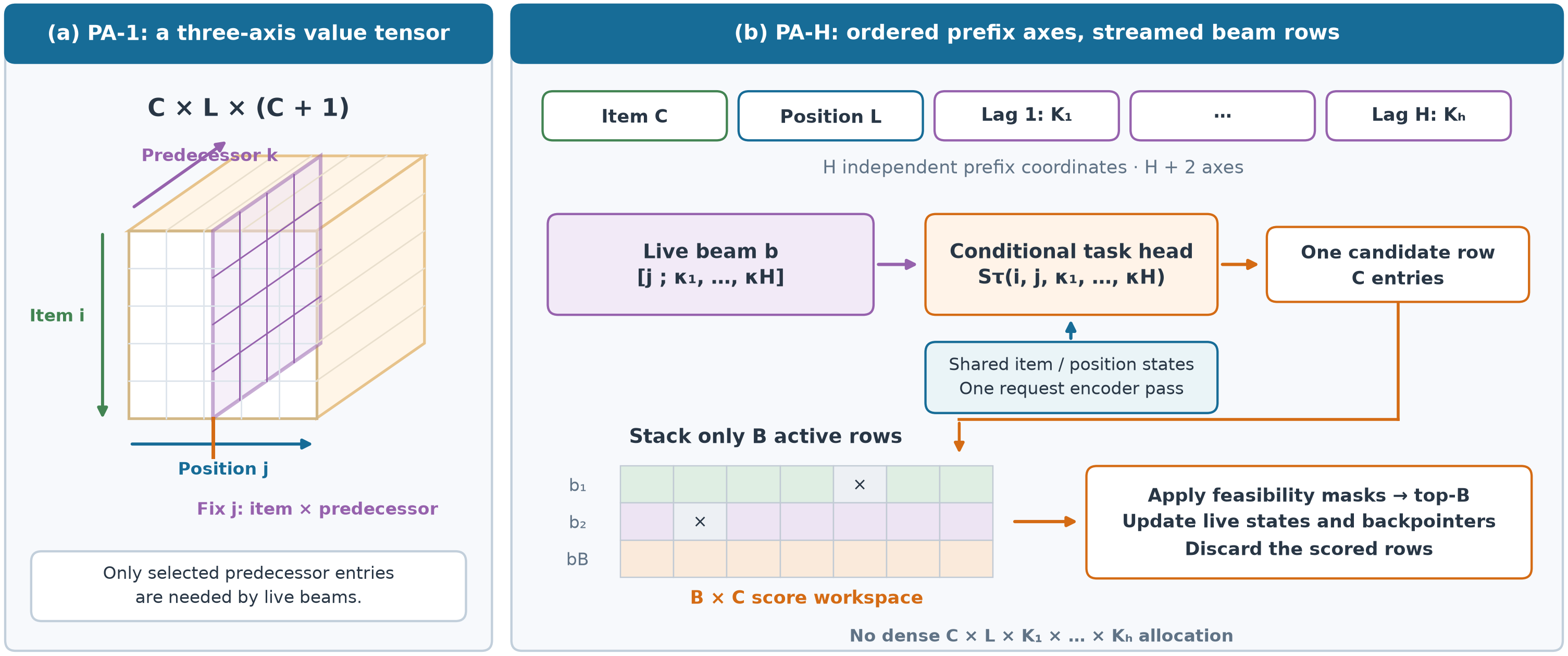}
\captionof{figure}{Implicit tensor indexing and active-row materialization.
(a) PA-1 has current-item, target-position, and predecessor axes.
Fixing a target position selects an item--predecessor plane; live beams
require only entries indexed by their selected predecessors.
(b) PA-$H$ retains $H$ independent ordered prefix coordinates, yielding
$H+2$ axes. Fixing position and a live prefix tuple yields a length-$C$
candidate row. Stacking the $B$ live rows forms a $B\times C$ workspace,
which is masked and discarded after top-$B$ selection.
No Cartesian product of prefixes is materialized.}
\Description{A PA-1 tensor highlights a fixed-position item-predecessor plane.
An ordered higher-order prefix tuple selects one candidate row per beam,
producing a small active-beam workspace.}
\label{fig:tensor-streaming}
\end{minipage}
\vspace{1em}
}]

\subsection{Active-row search equivalence and storage accounting}
\label{app:search-accounting}
Equation~\ref{eq:active-row} evaluates the same extension value as an
explicit tensor entry. For tuple-only PA-$H$, the retained coordinates
identify this value; a pooled-tail variant additionally retains its
aggregate statistic. Identical feasibility masks, cumulative scores, and
tie-breaking therefore yield identical next beams, inductively over
positions. This storage equivalence neither makes beam search globally
optimal nor makes compressed and exact histories the same scoring model.

The main storage bounds exclude weights and temporary encoder activations,
with task count and depth fixed. The evaluated PA-$N$ scorer recomputes the
prefix attention states from shared item representations per live beam,
rather than incrementally carrying attention states forward.
Optional dense PA-1 position slices require $O(C^2)$ scores instead of
$O(BC)$. Backpointers add $O(BL)$ entries, absorbed by $O(BC)$ for $L\leq C$.

\paragraph{Decoder profiling protocol.}
Full decoding uses one AMD EPYC-Genoa thread, ten warmups, and 100 timed
invocations with seeded weights and frozen catalog states. Beam width is
32; candidate counts are ten (KuaiRand) and nine (RL4RS).
The original scoring paths and beam functions are used, truncating only
the decoding loop and returned slate for length sweeps.
Timing includes lookup, prefix computation, conditional heads, uniqueness
masks, and top-$B$ selection; tower fitting and loading are excluded.
FLOPs count profiler-supported matrix products and elementwise
additions/multiplications, with two operations per multiply--accumulate.
Unsupported operators, lookup, masking, and sorting count toward latency
but not arithmetic; fused attention is expanded only for FLOP accounting.

\subsection{Quantizer fitting}
\label{app:quantizer-details}
Residual quantization assigns a scorer state $x_i$ to an ordered code tuple:
\begin{align}
 c_i^{(r)}&=\arg\min_a\|e_i^{(r-1)}-q_a^{(r)}\|_2^2,\\
 e_i^{(r)}&=e_i^{(r-1)}-\operatorname{sg}(q_{c_i^{(r)}}^{(r)}),\quad e_i^{(0)}=E_\eta(x_i),\\
 \hat z_i^{(r)}&=\sum_{s=1}^{r}q_{c_i^{(s)}}^{(s)}.
 \label{eq:rqvae}
\end{align}
A decoder reconstructs the item state from the selected levels.
For $R$ levels, define $z_i=E_\eta(x_i)$ and use stop-gradient
$\operatorname{sg}(\cdot)$ to form
$z_i^{\mathrm{st}}=z_i+\operatorname{sg}(\hat z_i^{(R)}-z_i)$.
The forward value is quantized, while gradients to the encoder pass through
$z_i$. A residual-VQ reconstruction objective is
\begin{equation}
 \begin{aligned}
 \mathcal{L}_{\mathrm{RQ}}={}&
 \sum_i\|x_i-D_\xi(z_i^{\mathrm{st}})\|_2^2\\
 &+\sum_{i,r}\bigl(
 \|\operatorname{sg}(e_i^{(r-1)})-q_{c_i^{(r)}}^{(r)}\|_2^2\\
 &\hspace{16mm}+\beta
 \|e_i^{(r-1)}-\operatorname{sg}(q_{c_i^{(r)}}^{(r)})\|_2^2\bigr).
 \end{aligned}
 \label{eq:rq-loss}
\end{equation}
Stop-gradient in the residual update leaves the greedy forward assignment
unchanged and separates codebook fitting from encoder commitment.
The residual terms fit the codebook and commit the encoder; $\beta>0$
controls commitment. Three 16-vector levels yield at most $4{,}096$ code tuples
from 48 primitives. A prefix axis has at
most $16^r+1$ states after including BOS; observed occupancy may be lower.
A decoder maps $\hat z_i^{(r)}$ back to the scorer's $d$-dimensional space.

\subsection{Evaluation protocols}
\label{app:evidence-boundaries}
Held-out AUC/NE use observed labels; public slate generation uses disjoint
order-sensitive evaluators; industrial data study uses the decoder's
scorer. Randomized scalability evaluation separately measures observed outcomes.
Compression uses frozen, context-independent tower states.

\section{Implementation and Aggregation Details}
\label{app:implementation}
\paragraph{Shared supervised representation.}
Prediction towers use 240,000 training windows; generator and evaluator
towers use disjoint 120,000-window halves. All are frozen before scoring
experiments. NAR/DCN/CGA/MIS share the resulting 32D states; DNN and
Seq2Slate fit separate towers with identical features and fold boundaries.
Generator fitting never uses evaluator labels or states.

\begin{table}[h]
\centering
\caption{Public scoring implementations. Counts exclude frozen catalog
embeddings and tower fitting. Counts are KuaiRand/RL4RS, whose slot
embedding sizes differ. All layers use width 32 unless stated otherwise.}
\label{tab:implementation}
\small
\setlength{\tabcolsep}{3pt}
\begin{tabular}{llr}
\toprule
Model & Scoring architecture & Parameters \\
\midrule
DNN & $32\!\to\!1\!\to\!1$ MLP & 35/35 \\
DCN & 3 cross layers + MLP & 8,097/8,065 \\
NAR & Item--slot inner product & 320/288 \\
MIS--PA-1 & $96\!\to\!32\!\to\!1$ residual & 4,513/4,481 \\
CGA & 2 causal layers + scalar head & 18,497/18,465 \\
MIS--PA-$N$ & 2 causal layers + attention & 29,857/29,825 \\
Seq2Slate & GRU encoder/pointer + head & 15,873/15,873 \\
\bottomrule
\end{tabular}
\end{table}

\paragraph{Optimization and objectives.}
Scorers use binary cross-entropy, Adam (0.003), batches of 512, 12 epochs,
and seeds 0--4, reporting final-epoch statistics. CGA/PA-$N$ have two
width-32 Transformer layers, four heads, feed-forward width 64, and no
dropout. Public PA-$N$ shares one position table between lag states
$e_\ell$ and target slots $t_j$, with a candidate-query projection and
$128\!\to\!64\!\to\!32\!\to\!1$ residual initialized to zero at its last layer.
DNN has a one-dimensional bottleneck.
Seq2Slate uses a 32D GRU encoder, GRU-cell decoder, dot-product pointer,
and separate binary outcome head. Its loss combines position-averaged
permutation cross-entropy and outcome cross-entropy. The target permutation
sorts logged binary relevance with stable ties; AUC/NE use the outcome
head, while generation sums pointer log-probabilities.

\paragraph{Seeds, beams, and evaluator scores.}
AUC/NE average per-seed metrics. NAR/MIS average five members' sigmoid
extension scores before beam selection; CGA/Seq2Slate decode each member
and average evaluator values. All reported generation uses beam 32 and
the first 2,000 held-out slates.
The main evaluator is a five-member causal Transformer with its own
tower, fitted only on the evaluator half. Seed 21000 fixes a random
reference permutation per slate. For generated value $v_s$ and reference
$r_s$, lift is
\[
100\,\frac{\frac{1}{S}\sum_s(v_s-r_s)}
{\left|\frac{1}{S}\sum_s r_s\right|}.
\]
Intervals use 2,000 paired slate bootstrap samples.
Seq2Slate fits separate towers/evaluator under the same architecture and
fold boundaries, with random-reference seed 29000.

\paragraph{Additional timing detail.}
Mean-latency intervals are $\bar t\pm1.96s_t/\sqrt{100}$.
MIS/CGA intervals are $[6.746,6.781]$/$[27.454,27.631]$ ms on
KuaiRand and $[5.684,5.758]$/$[20.237,20.323]$ ms on RL4RS.
The archived five-member MIS ensemble takes 30.820/25.746 ms;
Table~\ref{tab:matched-results} reports single members.
The remaining-decoder benchmark uses the original NAR/PA-1 paths at
$B=32$, $L=C$, with parameter counts in Table~\ref{tab:implementation}.

\paragraph{Industrial prefix implementation.}
The industrial-scale
MIS model uses PA-4: it retains the immediately preceding item embedding and
encodes content type and relative position for four recent lags:
\begin{align}
 h_m &= e_{\mathrm{type}(\kappa_{j,m})}+e_{\mathrm{relpos}(m)},\\
 P_j &= \operatorname{MLP}([W_x x_{\kappa_{j,1}};h_4;h_3;h_2;h_1]).
 \label{eq:pa4}
\end{align}
BOS supplies missing lags. This type-based model is separate from
the public PA-4 + pooled-tail model and full-prefix MIS--PA-$N$.
Rank-based input positional encodings are disabled on the item side;
selected-prefix lag embeddings and output-slot context remain available. MIS denotes this PA-4 model throughout the industrial and
online studies; MIS-PA1 provides the first-order comparison.

\paragraph{Industrial scale comparison.}
The video-feed readout compares MIS with an industrial ranker control
using one-layer hash assignment and 300 segments per arm. Launch can require
realization improvement without guardrail regression.
Serving uses seven consecutive days over these arms without the realization
A/A filter. The realization metric excludes A/A traffic; engagement uses its aggregate
query population for overall video consumption and app-level linked sessions.

\end{document}